\documentclass[runningheads]{llncs}
\usepackage{graphicx}

\usepackage{tikz}
\usepackage{comment}
\usepackage{amsmath,amssymb} 
\usepackage{color}
\usepackage{cite}
\usepackage{multirow}
\usepackage{pgfplots}
\usepackage{caption}
\usepackage{subcaption}

\usepackage[accsupp]{axessibility}  

\usepackage[width=122mm,left=12mm,paperwidth=146mm,height=193mm,top=12mm,paperheight=217mm]{geometry}

\begin{document}
\pagestyle{headings}
\mainmatter
\def\ECCVSubNumber{11}  

\title{Out-of-Vocabulary Challenge Report} 

\titlerunning{Out-of-Vocabulary Challenge Report}
%
\author{
Sergi Garcia-Bordils\textsuperscript{$\dagger$}\textsuperscript{1,3}\and
Andrés Mafla\textsuperscript{$\dagger$}\inst{1} \and
Ali Furkan Biten\textsuperscript{$\dagger$}\inst{1} \and \\
Oren Nuriel\textsuperscript{*}\inst{2} \and
Aviad Aberdam\textsuperscript{*}\inst{2} \and
Shai Mazor\textsuperscript{*}\inst{2} \and
Ron Litman\textsuperscript{*}\inst{2} \and \\
Dimosthenis Karatzas\inst{1}
}
\authorrunning{S. Bordils et al.}
%
\institute{
Computer Vision Center, Universitat Autonoma de Barcelona, Barcelona, Spain \and AWS AI Labs \and AllRead MLT\\
\email{\{sgbordils,amafla,abiten,dimos\}@cvc.uab.es}, \\
\email{\{onuriel,aaberdam,smazor,litmanr\}@amazon.com}
}
\maketitle

\begingroup\renewcommand\thefootnote{$\dagger$}
\footnotetext{Equal contribution}
\endgroup

\begingroup\renewcommand\thefootnote{$*$}
\footnotetext{Work does not relate to Amazon position}
\endgroup

\begin{abstract}
This paper presents final results of the Out-Of-Vocabulary 2022 (OOV) challenge. The OOV contest introduces an important aspect that is not commonly studied by Optical Character Recognition (OCR) models, namely, the recognition of unseen scene text instances at training time. The competition compiles a collection of public scene text datasets comprising of 326,385 images with 4,864,405 scene text instances, thus covering a wide range of data distributions. A new and independent validation and test set is formed with scene text instances that are out of vocabulary at training time. The competition was structured in two tasks, end-to-end and cropped scene text recognition respectively. 
A thorough analysis of results from baselines and different participants is presented. Interestingly, current state-of-the-art models show a significant performance gap under the newly studied setting. We conclude that the OOV dataset proposed in this challenge will be an essential area to be explored in order to develop
scene text models that achieve more robust and generalized predictions.
\end{abstract}

\section{Introduction}

Scene-text detection and recognition plays a key role in a multitude of vision and language tasks, such as visual question answering~\cite{biten2019scene, singh2019towards}, image captioning~\cite{captioning} or image retrieval \cite{mafla2021multi}.
Performance on classic benchmarks, such as ICDAR13~\cite{karatzas2013icdar} or ICDAR15~\cite{karatzas2015icdar} has noticeably increased thanks to the surge of sophisticated deep learning models. Interest in this field has gained traction in the last few years and, as a consequence, multiple new datasets have appeared. Some of them have introduced diverse new challenges, such as irregular text detection and recognition~\cite{ch2017total, yao2012detecting} or complex layout analysis~\cite{long2022towards}. At the same time, the scale of new datasets has also noticeably increased, reducing the reliance on synthetic data \cite{singh2021textocr, krylov2021open}.

However, none of the existing benchmarks makes a distinction between out-of-vocabulary (OOV) words and in-vocabulary (IV) words. By OOV word we refer to text instances that have never been seen in the training sets of the most common Scene Text understanding datasets to date. Recent research suggests that current OCR systems over-rely on language priors to recognize text~\cite{wan2020vocabulary}, by exploiting their explicit or implicit language model. As a consequence, while the performance on IV text is high, recognition performance on unseen vocabulary is lower, showing poor generalization. Since OOV words can convey important high-level information about the scene (such as prices, dates, toponyms, URLs, etc.), performance of OCR systems on unseen vocabulary should also be seen as an important characteristic.

With this motivation in mind we present the Out-of-Vocabulary Challenge, a competition on scene text understanding where the focus is put on unseen vocabulary words. This challenge is formed by two different tasks; an End-to-End Text Recognition task and a Cropped Word Text Recognition task. In the End-To-End task participants were provided with images and were expected to localize and recognize all the text in the image at word granularity. In the Cropped Word task the participants were presented with the cropped word instances of the test set, and were asked to provide a transcription. In order to be able to compare the performance of the submissions on seen and unseen vocabulary, we decided to include both types of instances on the test sets and report the results separately.

The dataset used for this competition is a collection of multiple existing datasets. Some of the featured datasets were collected with text in mind, while others used sources where the text is incidental.
We have created our own validation and test splits of the End-to-End challenge to contain at least one OOV word per  image. In the validation and test sets of the Cropped Word Recognition task we include the cropped OOV and IV words from the End-to-End dataset.

\section{Related Work}
The field of scene text recognition can be divided into two main tasks - text detection and text recognition. The OOV challenge addresses methods which either perform end-to-end text recognition \cite{li2017towards, liu2020abcnet,qin2019towards, feng2019textdragon, baek2020character, zhang2022text, lyu2018mask, kittenplon2022towards, ronen2022glass} and thus solve both tasks, or methods that tackle just text recognition \cite{shi2016end,baek2019wrong,litman2020scatter,slossberg2020calibration,qiao2020seed,wang2021two,nuriel2021textadain,seqclr_aberdam2021sequence,abinet_fang2021read,aberdam2022multimodal}, and thus assume the words are already extracted. 

The problem of vocabulary reliance in scene text recognition was first revealed by \cite{wan2020vocabulary}. Through extensive experiments, they found that state-of-the-art methods perform well on previously seen, in vocabulary, word images yet generalize poorly to images with out of vocabulary words, never seen during training. In addition, the authors proposed a mutual learning approach which jointly optimizes two different decoder types, showing that this can alleviate some of the problems of vocabulary reliance. \cite{zhang2022context} suggested a context-based supervised contrastive learning framework, which pulls together clusters of identical characters within various contexts and pushes apart clusters of different characters in an embedding space. In this way they are able to mitigate some of the gaps between in and out of vocabulary words.

\section{Competition Protocol}

The OOV Challenge took place from May to July of 2022.
A training set was given to participants at the beginning of May, but the images for the test set were only made accessible for a window between June 15 and July 22.
Participants were asked to submit results obtained on the public test set images rather than model executables.
We rest on the scientific integrity of the participants to adhere to the challenge's specified guidelines. 

The Robust Reading Competition (RRC) portal\footnote{\url{https://rrc.cvc.uab.es/}}  served as the Challenge's host.
The RRC site was created in 2011 to host the first rigorous reading contests including text detection and identification from scene photographs and born-digital images, and it has since expanded into a fully-fledged platform for organizing academic competitions. 
The portal now hosts 19 distinct challenges, with different tasks mostly related to scene text detection and recognition, Scene-Text Visual Question Answering (ST-VQA) and Document VQA.
The RRC portal has more than 35,000 registered users from more than 148 countries, and more than 77,000 submitted results have already been evaluated.
The findings in this report are an accurate reflection of the submissions' status at the end of the formal challenge period.
The RRC portal should be viewed as an archive of results, where any new findings contributed after the compilation of this report will also be included.
All submitted findings are automatically analyzed, and the site provides per-task ranking tables and visualization tools to examine the results. 

\section{The OOV Dataset}

The OOV (Out Of Vocabulary) dataset encompasses a collection of images from 7 public datasets, namely: HierText~\cite{long2022towards}, TextOCR~\cite{singh2021textocr}, ICDAR13~\cite{karatzas2013icdar}, ICDAR15~\cite{karatzas2015icdar}, MLT19~\cite{nayef2019icdar2019}, Coco-text~\cite{veit2016coco} and OpenImages~\cite{krylov2021open}. This dataset selection aims to generalize the performance of models overcoming existing biases in datasets~\cite{khosla2012undoing}.

The validation and test splits of the OOV dataset were defined to measure the performance of models on unseen words at training time. To do this we extracted all the words that appear at least once in the training and validation splits of the datasets and, jointly with the 90k word dictionary introduced by Jaderberg et al.~\cite{jaderberg2014synthetic}, we created an in-vocabulary dictionary of words. To create the test set of the OOV dataset we picked those images from the original test sets which contained, at least, one word outside of this vocabulary. Images in the validation dataset were picked from the training and validation splits of the original datasets, and we only kept images that contained words that appear once (and therefore do not appear in the training split). We limited the number of images in the validation set to 5,000 images. The rest of the images were used in the training split. 

For this first iteration of the competition we focus on text instances in which characters come from a limited alphabet. This alphabet is formed by the Latin alphabet, numbers and a few punctuation signs \footnote{See \url{https://rrc.cvc.uab.es/?ch=19\&com=tasks} for the full alphabet.}. Words that contain out of alphabet characters are not considered for the final evaluation (in the End-to-End task they are treated as ``don't care''). 

\begin{table}[t!]
\centering
\footnotesize
    \begin{tabular}{l|lll|lll}
        & \multicolumn{3}{c|}{\textbf{\# of Images}} & \multicolumn{3}{c}{\textbf{\# of Cropped Words}} \\
        \hline
        Dataset & Train & Validation & Test & Train & Validation & Test \\ 
        \hline
        ICDAR13 & 229 & 0 & 233 & 795 & 54 & 0 \\
        ICDAR15 & 1000 & 0 & 500 & 4350 & 118 & 0\\
        ICDAR MLT19 & 10000 & 0 & 10000 & 80937 & 8499 & 103297 \\ 
        MSRA-TD500 & 300 & 0 & 200 & - & - & -\\
        COCO-Text & 43686 & 10000 & 10000 & 80549 & 6571 & 11688 \\
        TextOCR & 24902 & 0 & 3232 & 1155320 & 47019 & 96082 \\ 
        HierText & 8281 & 1724 & 1634 & 981329 & 59630 & 187190 \\ 
        Open Images V5 Text & 191059 & 16731 & 0 & 2066451 & 7788 & 0 \\
        \hline
        OOV Dataset & 312612 & 5000 & 8773 & 4369731 & 128832 & 365842 \\
        \hline
        \end{tabular}
    \caption{Dataset size comparison}
    \label{tab:e2e_stats}
    \vspace{-.9cm}
\end{table}
\begin{figure}
    \centering
    \includegraphics[width=\textwidth]{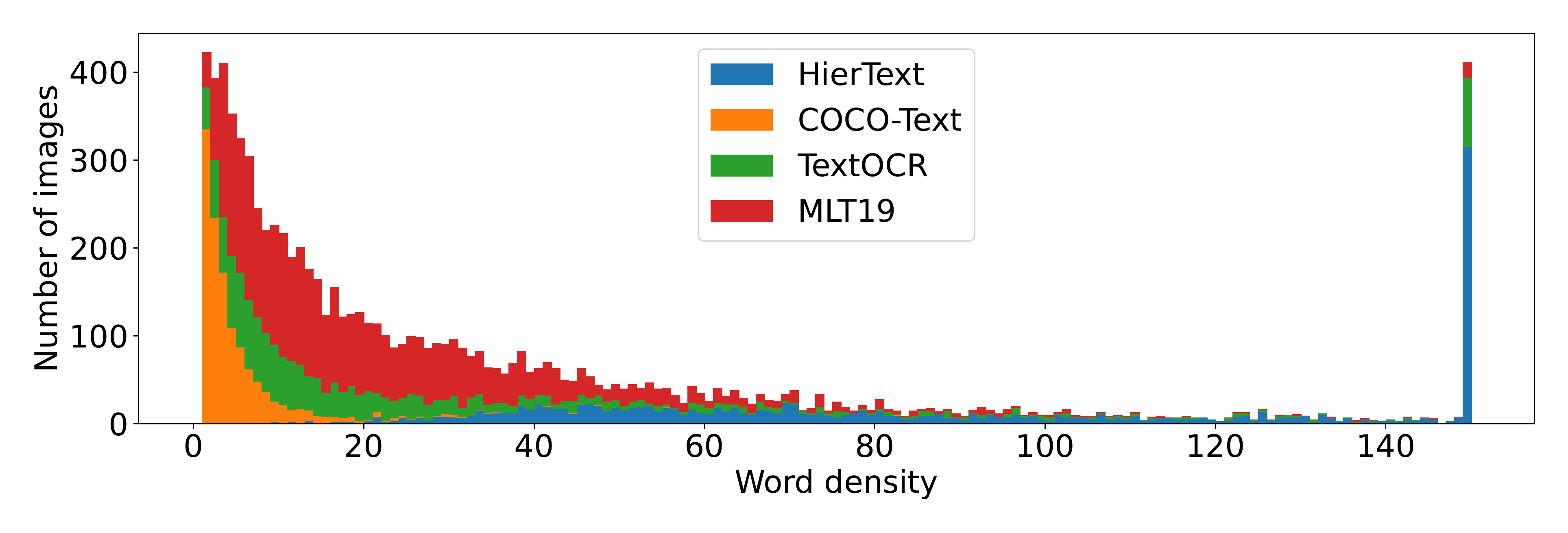}
    \caption{Number of words per image and dataset of origin. Images that contain more than 150 words have been counted in the last bin.}
    \label{fig:word_density}
    \vspace{-.9cm}
\end{figure}
\subsection{Dataset Analysis}  

Given that this dataset is a collection of datasets from multiple sources, it has the benefit of featuring data that come from different sources and annotation settings. Some of the featured datasets were originally collected with text in mind, while in others the text is more incidental. COCO-Text is an example of a dataset with purely incidental text. Since the source of the images is the MS COCO~\cite{lin2014microsoft} dataset, the images are not text biased like in other datasets. In Figure \ref{fig:word_density} we can see the distribution of the number of words per image and dataset of origin in the test set. Some datasets contain, in general, a small number of instances per image (such as the aforementioned COCO-Text), while others are more prone to contain images with tens or even hundreds of words (HierText being the most prominent one). Consequently, our test set is less biased towards specific distributions of words.

\begin{figure}[t!]
    \centering
    \captionsetup[subfigure]{labelformat=empty}
    
    \begin{subfigure}{.17\textwidth}
        \includegraphics[height=\textwidth, width=\textwidth]{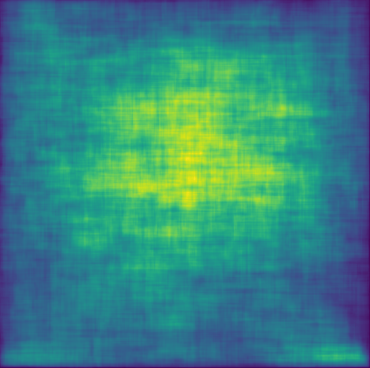}
        \caption{COCO-Text}
    \end{subfigure}
    \hfill
    \begin{subfigure}{.17\textwidth}
        \includegraphics[height=\textwidth, width=\textwidth]{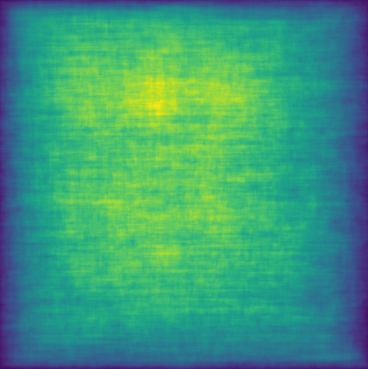}
        \caption{HierText}
    \end{subfigure}
    \hfill
    \begin{subfigure}{.17\textwidth}
        \includegraphics[height=\textwidth, width=\textwidth]{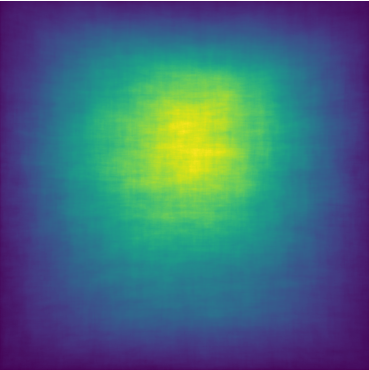}
        \caption{MLT19}
    \end{subfigure}
    \hfill
    \begin{subfigure}{.17\textwidth}
        \includegraphics[height=\textwidth, width=\textwidth]{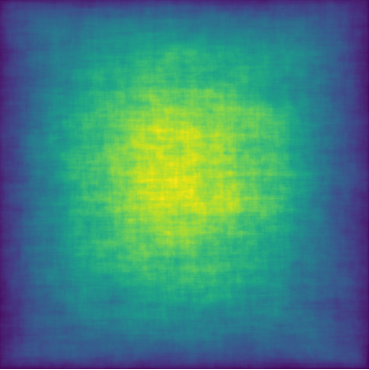}
        \caption{TextOCR}
    \end{subfigure}
    \hfill
    \begin{subfigure}{.17\textwidth}
        \includegraphics[height=\textwidth, width=\textwidth]{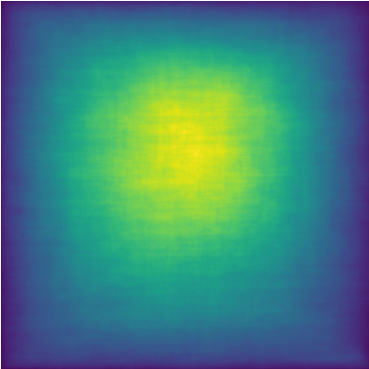}
        \caption{OOV}
    \end{subfigure}
    \caption{Text spatial distribution of the different datasets featured in the test set.}
    \label{fig:spatial_dist}
    \vspace{-.5cm}
\end{figure}

\begin{figure}[t!]
    \centering
    \includegraphics[width=.9\textwidth]{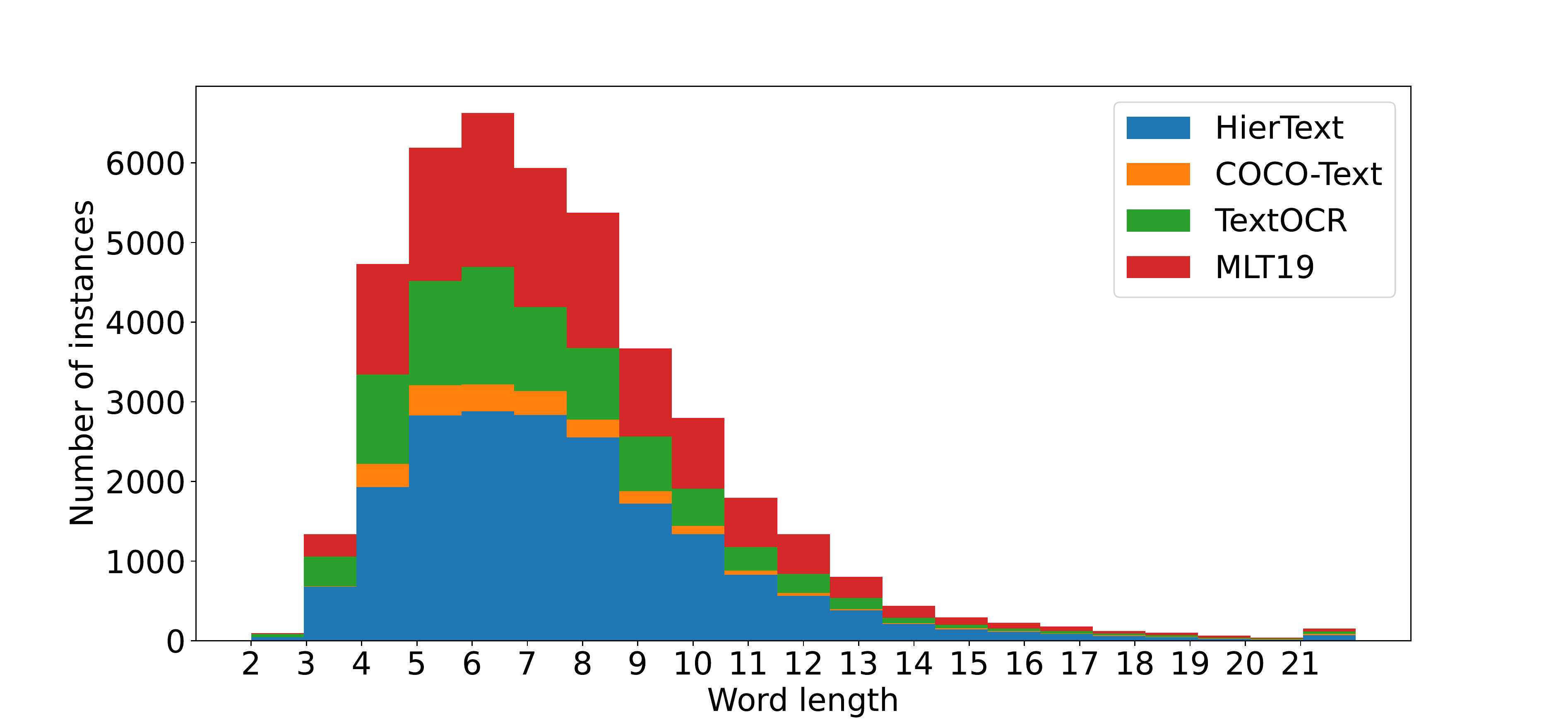}
    \caption{Character length of OOV words per dataset.}
    \label{fig:word_len_dist}
    \vspace{-.7cm}
\end{figure}

Table \ref{tab:crops_stats} shows the dataset of origin of the cropped words for each one of the splits. The test split features a balanced number of instances coming from 4 different sources, avoiding relying too much on a specific dataset. The spatial distribution of the instances of the test set can be seen in the Figure \ref{fig:spatial_dist}. Each dataset featured in the test set appears to have different spatial distributions, a consequence of the original source of the images.
For example, COCO-Text and TextOCR originally come from datasets that were not collected with text in mind (MS COCO~\cite{veit2016coco} and Open Images V4~\cite{kuznetsova2020open}), consequently the word instances are incidental and distributed over all the image. On the other hand, the text featured on MLT19 is more focused and more clustered around the center of the images. HierText was collected with text in mind but contains much more text instances per image (as seen in the Figure \ref{fig:word_density}, which distributes the text uniformly over the images. Combining datasets that contain images from different origins gives us a more varied and rich test set. Finally, Figure \ref{fig:word_len_dist} shows the distribution of the lengths of the OOV words of the test set, separated per dataset. The distribution of the lengths appears to be similar for the featured datasets.

\section{The OOV Challenge}

The Out of Vocabulary challenge aims to evaluate the ability of text extraction models to deal with words that they have never seen before. Our motivation for organizing this challenge is the apparent over-reliance of modern OCR systems on previously seen vocabulary. We argue that generalizing well on in and out of vocabulary text should be considered as important as generalizing well on text with different visual appearances or fonts. OOV text can convey important semantic information about the scene, failing to recognize a proper noun (such as the name of a street in the context of autonomous driving) or a random string of numbers (such as a telephone number)
can result in unfortunate consequences.

Wan \textit{et. al.} \cite{wan2020vocabulary} call the phenomenon of memorizing the words the in the training set \textit{``vocabulary reliance''}. To prove this behaviour, the authors train diverse text recognition models using the same data and the same backbone. The results are reported using the IIIT-5k\cite{mishra2012scene} dataset and they provide results on both in-vocabulary and out-of-vocabulary text. Like us, they consider OOV whichever words are not present in the test set, including the synthetic data. Their results show a gap as small as 15.3\% using CA-FCN~\cite{liao2019scene} and as high as 22.5\% using CRNN~\cite{shi2016end}, proving how much OCR systems rely on learned vocabulary.

Therefore, in this challenge we evaluate the entries putting special emphasis on words that the models have never seen before. We hope that our curated dataset and our evaluation protocol can be useful to the community to develop more robust and unbiased OCR systems.

\subsection{Task 1}

The End-to-End task aims to evaluate the performance of the models in both detection and recognition. Unlike some previous competitions in the RRC, we do not provide any vocabulary. For each correct detection we look for a perfect match between the proposed and the ground truth transcriptions.
Evaluation is case-sensitive, and punctuation signs are taken into account.
The evaluation procedure displays results on both IV and OOV text for each of the methods. Below, we give brief descriptions for some of the submitted methods, as provided by their authors.

\noindent
\textbf{CLOVA OCR DEER.} An end-to-end scene text spotter based on a CNN backbone, a Deformable Transformer Encoder~\cite{zhu2020deformable}, location decoder and text decoder. The location decoder, based on the segmentation method (Differentiable Binarization~\cite{liao2019scene}), detects text regions, and the text decoder based on the deformable transformer decoder recognizes each instance from image features and detected location information.
They use both the training data provided by the challenge, as well as synthetic data. 

\noindent
\textbf{Detector Free E2E.} 
It is a detection free end-to-end text recognizer, where a CNN with Deformable Encoder \& Decoder is used. The models are trained with data from the challenge and additional SynthText data synthesized with MJSynth 90k dictionary.

\noindent
\textbf{oCLIP \& oCLIP\_v2.} For detection, they first pre-train their Deformable ResNet-101 by using oCLIP on the provided training set. Then they train TESTR~\cite{zhang2022text}, PAN and Mask TextSpotter with different backbones using the pre-trained model. Finally, they combine results from different methods, different backbones, and different scales together while for recognition, they adopt SCATTER~\cite{litman2020scatter}. 


\noindent
\textbf{DB\_threshold2\_TRBA.} The detector is based on Differentiable Binarization (DB)~\cite{liao2020real}. The recognizer is TRBA from WIW~\cite{baek2019wrong}.
TRBA denotes TPS + ResNet Backbone + BiLSTM + Attention. The models were not jointly trained. 
Since DB does not output an up-vector, they rotated the detected region according to the aspect ratio. CocoText has label noises (not case sensitive), and thus, they cleaned the dataset using the teacher model. They use synthetic data (ST) as well as challenge-provided data.

\noindent
\textbf{E2E\_Mask.} They only use the OOV dataset to train their model. In the detection stage, they follow TBNet and Mask2Former as the base model with a multi-scale training strategy.
To combine the final detection results, they ensemble different detectors with different backbones and different testing sizes. In the recognition stage, they use a vision transformer model that consists of ViT encoder and query-based decoder to generate the recognition results in parallel. 




\subsection{Task 2}

On the Cropped Word Recognition task the participants had to predict the recognition for all the cropped words of the test set. There is a total of 313,751 cropped words in the test set, 271,664 of these words are in-vocabulary and 42,087 are out of vocabulary. Like in Task 1, 
the evaluation is case-sensitive, and punctuation signs are taken into account.
Below, we provide a brief description for some of the submitted methods.

\noindent
\textbf{OCRFLY\_V2.} They design a new text recognition framework for OOV-ST, named Character level Adaptive Mutual Decoder (CAMD), where both multi-arch and multi-direction autoregressive seq2seq heads are jointly used during training and testing. CAMD adopts a CNN-ViT Backbone as encoder, and two different vision-language adaptively balanced decoders: an LSTM and a Transformer decoder, are built upon the aforementioned encoder. Only Syn90k and the training splits given by the challenge are used for training.

\noindent
\textbf{OOV3decode.} Three models are combined by voting. An Encoder-Decoder Framework with a 12-ViT-based encoder, a CTC Combined Decoder, a CTC-Attention Combined Decoder, and a Mix-CTC-Position-Attn Combined Decoder.
They use 300w+ generated images and the challenge training data.

\noindent
\textbf{Vision Transformer Based Method (VTBM).} They train several models with the same Vision Transformer based backbone and various decoders (CTC and Attention), and they ensemble them based on confidence.
They first pre-trained their models on nearly 10 million synthetic images and fine-tuned them on the official training set.
Common augmentations such as rotation, blur, etc. are adopted; especially the image concatenate augmentation is used to mine textual context information.

\noindent
\textbf{DAT.} An Encoder-Decoder transformer-based encoder with 12 layers of VIT-based block and $4\times4$ patch size is used.
An ensemble strategy is used to fuse the results from three decoder types: a CTC-based decoder, an attention-based decoder and a CTC+attention-based decoder.

\noindent
\textbf{OCRFLY.} This is a simple baseline based on the seq2seq algorithm, where they adopt a CNN-ViT Backbone as encoder and a 6-layer transformer as decoder. Only Syn90k and the training splits are used for training.

\subsection{Baselines}
For Task 1, we evaluate the recent state of the art methods, TESTR \cite{zhang2022text}, TextTranSpotter\cite{kittenplon2022towards}, GLASS\cite{ronen2022glass}.

\noindent
\textbf{TESTR.} We provide results for their pretrained model released on their official code package\footnote{\url{https://github.com/mlpc-ucsd/testr}} using the default configurations.

\noindent
\textbf{TextTranSpotter.} TextTranSpotter was trained following the fully-supervised training protocol in the paper with the following datasets: pretrained on SynthText, then fine-tuned on a mix of SynthText, ICDAR13, ICDAR15, TotalText and TextOCR. The model weights were then frozen and the mask branch was trained on SynthText and then on the mix of datasets. 

\noindent
\textbf{GLASS.} Training was performed on the following train datasets: SynthText, ICDAR13, ICDAR15, TotalText and TextOCR. The model was pretrained for 250k iterations with a batch size of 24, and then fine tuned for another 100k iterations specifically on TextOCR with a batch size of 8 images.
The architecture and parameters chosen for the detection, recognition and fusion branches, are detailed in \cite{ronen2022glass}.

For Task 2, we evaluate the models in two types of settings, the first when trained on synthetic data and the second when trained on the real data. The real data consists of the word crops introduced in the OOV dataset.

\noindent
\textbf{SCATTER.} In both settings SCATTER was trained for 600k iterations from scratch. We employ the exact same training procedure as described in the \cite{litman2020scatter}. We use two selective-contextual refinement blocks and take the output of the last one.

\noindent
\textbf{Baek et al.} For the synthetic setting, we use the case-sensitive model released in the official repository\footnote{\url{https://github.com/clovaai/deep-text-recognition-benchmark}}. For the real data setting the model was trained from scratch using the same training procedure as published in the repository.

\noindent
\textbf{ABINET.} We used the official codebase\footnote{\url{https://github.com/FangShancheng/ABINet}} and trained a case-sensitive model on the MJ and ST synthetic datasets for the synthetic setting. For the real setting, we also utilized the OOV dataset. In both cases, we only trained the network end-to-end without the pretraining stages.

\subsection{Evaluation Metrics}  

For Task 1 (End-to-End text detection and recognition) we use a modified version of the evaluation method proposed by Wang et. al.~\cite{wang2011end}. This method considers a correct match when one of the proposed detections overlaps with a ground truth bounding box by more than 50\% and their transcriptions match (again, caring about the letter case and punctuation signs). Correctly matched proposals count as true positives, while unmatched proposals count as false positives. Unmatched ground truth annotations count as false negatives. Most of the annotations of the datasets used to form the validation and test sets have annotations with some sort of ``unreadable'' attribute. These words are treated as ``don't care'', and do not affect (positively or negatively) the results. As discussed earlier, words that contain characters that are out of alphabet are also considered as ``don't care''. Jointly with the precision and recall, we also report the harmonic mean (or F-score) of each method:

\begin{equation}
Hmean = \frac{2 * Recall * Precision}{Recall + Precision}
\end{equation}

Since our evaluation protocol has to distinguish between OOV and IV words, we modified the evaluation procedure to ignore the opposite split during the evaluation. For example, when we are evaluating on OOV words, in-vocabulary words are treated as ``don't care''. This way, matched ground truth IV words do not count as false positives, and unmatched words do not count as false negatives. The opposite applies when evaluating for IV words. The number of false positives is the same in both cases, since matched ground truth annotations are either true positives or treated as ``don't care''.

Finally, we report the unweighted average of the OOV and IV Hmean, as a balanced metric for this task. Our motivation behind this metric is to give equal importance to both distributions, as ideally, increasing performance on OOV words should not undermine performance on IV words.

For the Task 2 (Cropped Word Recognition) we report two metrics. The first metric is the total edit distance between each predicted word and its ground truth, 
considering equal costs for insertions, deletions and substitutions. 
The second metric is word accuracy, which is calculated as the sum of correctly recognized words divided by the total amount of text instances. We provide both metrics for OOV and IV words. The final, balanced metric reported for the Task 2 is the unweighted average of the OOV and IV accuracy. Similarly to Task 1, we consider performance on both subsets equally important.

\section{Results}
In this section, we provide the results for the submitted methods on both tasks. We also present baselines to compare to the submitted methods.

\begin{table}[t!]
\centering
    \resizebox{\textwidth}{!}{
    \begin{tabular}{lcccccccccc} 
        \hline
        \multirow{2}{*}{\textbf{Method}} & \textbf{Average}  & \multicolumn{3}{c}{\textbf{All}} & \multicolumn{3}{c}{\textbf{OOV}} & \multicolumn{3}{c}{\textbf{IV}}
      \\  & \textbf{Hmean} & \multicolumn{1}{c}{P} & \multicolumn{1}{c}{R} &  \multicolumn{1}{c}{Hmean} & \multicolumn{1}{c}{P} & \multicolumn{1}{c}{R} & \multicolumn{1}{c}{Hmean} &\multicolumn{1}{c}{P} & \multicolumn{1}{c}{R} & \multicolumn{1}{c}{Hmean}\\
        \hline
        TESTR\cite{zhang2022text}  & 15.9  & 31.4 & 20.1 & 25.1 & 4.4 & 17.8 & 7.1 & 29.2 & 21.4 & 24.6\\
        TextTranSpotter \cite{kittenplon2022towards}  & 18.6  & 37.4 & 25.0 & 29.9 & 4.5 & 16.4 & 7.0 & 35.5 & 26.2 & 30.1  \\
        GLASS\cite{ronen2022glass}  & 34.9  & 75.8 & 30.6 & 43.6 & 24.9 & 27.2 &  26.0 & 73.7 & 31.1 & 43.7 \\
  
        \hline
        CLOVA OCR DEER & \textbf{42.39} & \textbf{67.17} & \textbf{52.04} & \textbf{58.64} & 18.58 & 48.72 & 26.9 & \textbf{64.51} & \textbf{52.49} & \textbf{57.88} \\
        
        Detector Free E2E & 42.01 & 66.15 & 52.44 & 58.5 & 17.97 & 49.35 & 26.35 & 63.44 & 52.86 & 57.67 \\
        
        oCLIP\_v2 & 41.33 & 67.37 & 46.82 & 55.24 & 20.28 & 48.42 & 28.59 & 64.41 & 46.6 & 54.08 \\
        
        DB threshold2 TRBA & 39.1 & 64.08 & 49.93 & 56.13 & 15.26 & 42.29 & 22.43 & 61.6 & 50.96 & 55.78 \\
        
        E2E\_Mask & 32.13 & 47.9 & 54.14 & 50.83 & 8.64 & 46.73 & 14.58 & 45.2 & 55.14 & 49.68 \\
        
        YYDS & 28.68 & 51.53 & 35.54 & 42.07 & 10.63 & 33.36 & 16.12 & 48.57 & 35.83 & 41.24 \\ 
        
        sudokill-9 & 28.34 & 51.62 & 34.08 & 41.06 & 11.03 & 33.22 & 16.56 & 48.54 & 34.2 & 40.12 \\
        
        PAN & 28.13 & 50.5 & 34.81 & 41.21 & 10.5 & 33.58 & 16.0 & 47.45 & 34.98 & 40.27 \\
        
        oCLIP & 24.04 & 47.72 & 7.51 & 12.98 & \textbf{41.21 }& \textbf{48.42 }& \textbf{44.52} & 17.46 & 1.98 & 3.55 \\ 
        
        DBNetpp & 20.34 & 39.42 & 27.0 & 32.05 & 5.62 & 20.75 & 8.85 & 37.15 & 27.84 & 31.83 \\
        
        TH-DL & 9.32 & 18.39 & 13.23 & 15.39 & 2.16 & 10.87 & 3.6 & 16.89 & 13.55 & 15.04 \\
        
        \hline
        \end{tabular}}
    \caption{Harmonic mean, precision and recall for the entire dataset (All), in vocabulary (IV) and out of vocabulary (OOV) across different baseline methods. The average Hmean is the unweighted average of IV and OOV.}
    \label{tab:e2e_performance}
    \vspace{-.9cm}
\end{table}
\subsection{Task 1}
We provide all the results in Table~\ref{tab:e2e_performance} where the top part represents the baselines while the bottom part is for the submitted methods. We present results on two different subsets, namely in-vocabulary (IV) and out-of-vocabulary (OOV). Moreover, we rank the methods according to the balanced Average Hmean metric. 


As can be appreciated from Table~\ref{tab:e2e_performance}, CLOVA OCR DEER is the winning method in terms of Average Hmean. The best method in terms of OOV Hmean is oCLIP, surpassing the second best method by \textbf{+17.6} points. However, we see that oCLIP's performance on IV set is the lowest between the participated methods. This observation makes us wonder if there is a trade-off between the performance of IV and OOV in the model.

\begin{table}[t!]
    \centering
    \begin{tabular}{lcccccc} 
        \hline
        \multirow{2}{*}{\textbf{Method}} & \multirow{2}{*}{\textbf{Train Set}}  & \textbf{Total} & \multicolumn{2}{c}{\textbf{IV}} & \multicolumn{2}{c}{\textbf{OOV}}
      \\  &  & \multicolumn{1}{c}{Word Acc$\uparrow$} & \multicolumn{1}{c}{Word Acc$\uparrow$} & \multicolumn{1}{c}{ED$\downarrow$} & \multicolumn{1}{c}{Word Acc$\uparrow$} & \multicolumn{1}{c}{ED$\downarrow$} \\
        \hline
        ABINet \cite{abinet_fang2021read} & Syn & 38.01 & 50.29 & 342,552 & 25.73 & 115536 \\
        Baek et al \cite{baek2019wrong} & Syn & 44.47 & 52.61 & 365,566 & 36.34 & 114,101 \\
        SCATTER \cite{litman2020scatter} & Syn & 47.79 & 56.85 & 321,101 & 38.74 &  103,928  \\
        ABINet \cite{abinet_fang2021read} & Real & 59.84 & 71.13 & 176,126 & 48.55 & 67478\\
        Baek et al \cite{baek2019wrong} & Real & 64.97 & 75.98 & 138,479 & 53.96 & 54,346  \\
        SCATTER \cite{litman2020scatter} & Real & 66.68 & 77.98 & 128,219 & 55.38 & 52,535 \\
        \hline
        
        OCRFLY\_v2 & Syn + Real & \textbf{70.31} & 81.02 & 123,947 & \textbf{59.61} & 46,048 \\
        
        OOV3decode & Syn + Real & 70.22 & \textbf{81.58} & \textbf{94,259} & 58.86 & 40,175 \\ 
        
        VTBM & Syn + Real & 70.00 & 81.36 & 94,701 & 58.64 & 40,187 \\ 
        
        DAT & - & 69.90 & 80.78 & 96,513 & 59.03 & \textbf{40,082} \\
        
        OCRFLY & Syn + Real & 69.83 & 80.63 & 131,232 & 59.03 & 53,243 \\ 
        
        GGUI & - & 69.80 & 80.74 & 96,597 & 58.86 & 40,171 \\ 
        
        vitE3DCV & Syn + Real & 69.74 & 80.74 & 96,477 & 58.74 & 40,115 \\ 
        
        DataMatters & Syn + Real & 69.68 & 80.71 & 96,544 & 58.65 & 40,177 \\ 
        
        MaskOCR & Real & 69.63 & 80.60 & 108,894 & 58.65 & 44,971 \\ 
        
        SCATTER & Syn + Real & 69.58 & 79.72 & 113,482 & 59.45 & 43,89 \\ 
        
        Summer & Syn + Real & 68.77 & 79.48 & 103,211 & 58.06 & 42,118 \\ 
        
        LMSS & Syn + Real & 68.46 & 80.81 & 116,503 & 56.11 & 51,165 \\ 
        
        UORD & Real & 68.28 & 79.28 & 118,185 & 57.27 & 48,517 \\
        
        PTVIT & Syn + Real & 66.29 & 77.52 & 120,449 & 55.06 & 49,41\\   
        
        GORDON & Syn + Real & 65.86 & 77.25 & 124,347 & 54.47 & 48,907 \\ 
        
        TRBA\_CocoValid & Syn + Real & 63.98 & 77.76 & 132,781 & 50.20 & 60,693 \\ 
        
        HuiGuan & Real & 63.73 & 74.77 & 162,87 & 52.69 & 68,926 \\
        
        EOCR & - & 46.66 & 55.30 & 350,166 & 38.02 & 113,317 \\ 
        
        NNRC & - & 38.54 & 45.36 & 405,603 & 31.73 & 136,384 \\
        
        NN & - & 37.17 & 43.38 & 426,074 & 30.97 & 144,032 \\ 
        
        CCL & Real & 31.06 & 47.40 & 552,57 & 14.73 & 202,087 \\ 
        \hline
        \end{tabular}
    \caption{Word accuracy and Edit Distance for state of the art recognition models trained on different datasets.}
    \label{tab:recog_performance}
    \vspace{-1.1cm}
\end{table}

Regarding the architecture choices of the submitted methods, almost all of them, especially the top performing ones make use of the ViT~\cite{dosovitskiy2020image} architecture either as backbone for the recognition pipeline or directly for extracting features. Another commonly preferred building block is CTC~\cite{graves2006connectionist} based encoder or decoder mechanism. Furthermore, the top 2 performing methods, CLOVA OCR DEER and Detector Free E2E, utilize a Deformable DETR architecture~\cite{zhu2020deformable}, showing its effectiveness in the end-to-end text recognition task. Lastly, we observe that almost all the methods make use of synthetic data to either pre-train or finetune together with the real data.

\subsection{Task 2}

The results can be found in Table~\ref{tab:recog_performance} where the top part represents the baselines while the bottom part is for the submitted methods. We present results exactly the same way as in Task 1 in two different subsets, namely in-vocabulary (IV) and out-of-vocabulary (OOV) words. Moreover, we rank the methods according to the balanced Total Word Accuracy which is calculated as the average of the Word Accuracy of IV and OOV, giving the same emphasis on both sets. 
As can be seen from Table~\ref{tab:recog_performance}, our baselines trained with real data clearly outperform the ones trained with synthetic data. We also observe in our baselines that having a boost in IV words also translates to an improvement in OOV performance. 

Regarding the submitted methods, the winner in total word accuracy is OCRFLY\_v2 even though by a slight margin. We also notice that OCRFLY\_v2 is the best method in OOV performance in terms of accuracy; however, DAT is the best method in terms of edit distance. On the other hand, OOV3decode achieves state of the art performance in IV. 
We note that all top 3 methods are trained with both synthetic and real data. As a matter of fact, most of the methods are trained with combined data, confirming the effectiveness of the usage of the combined data. In terms of the favored architectures, we see a similar trend in Task 2 as in Task 1. We observe that the ViT~\cite{dosovitskiy2020image} architecture being used in most of the methods combined with CTC and Attention. This demonstrates the clear advantage of the Transformer architecture over LSTMs and RNNs which was state-of-the-art in text recognition literature previously.

\section{Analysis}

\begin{figure}[t!]
    \centering

    \begin{tikzpicture}
        \begin{axis}[
            width=.5\textwidth,
            height=0.3\textwidth,
            xlabel=Word length,
            ylabel=Average recall,
            mark size=1.5pt,
            legend style={font=\tiny},
            tick label style={font=\tiny},
            label style={font=\tiny},
            ytick={0.1,0.3,0.5,0.7,0.9},
            xtick={0,5,10,15,20,25},
            title={Detector Free E2E}
        ]
        \addplot coordinates{
            (2.0, 0.125)
            (3.0, 0.3639760837070254)
            (4.0, 0.4234995773457312)
            (5.0, 0.4622595765314369)
            (6.0, 0.4932106216053108)
            (7.0, 0.5101937657961246)
            (8.0, 0.5392631187197618)
            (9.0, 0.5390901661672569)
            (10.0, 0.5385989992852037)
            (11.0, 0.5261692650334076)
            (12.0, 0.531039640987285)
            (13.0, 0.5286783042394015)
            (14.0, 0.5194508009153318)
            (15.0, 0.48135593220338985)
            (16.0, 0.5265486725663717)
            (17.0, 0.4748603351955307)
            (18.0, 0.4959349593495935)
            (19.0, 0.5192307692307693)
            (20.0, 0.4461538461538462)
            (21.0, 0.5384615384615384)
            (22.0, 0.2857142857142857)
            (23.0, 0.2777777777777778)
            (24.0, 0.5384615384615384)
            (25.0, 0.14754098360655737)
        };
        \addplot coordinates {
            (2.0, 0.5947362008041915)
            (3.0, 0.3567312860124739)
            (4.0, 0.6857107933015524)
            (5.0, 0.7192566891305798)
            (6.0, 0.7354769097971515)
            (7.0, 0.7348277747402953)
            (8.0, 0.7433607145705399)
            (9.0, 0.7324303603373371)
            (10.0, 0.7330351934668481)
            (11.0, 0.689772341148488)
            (12.0, 0.6820809248554913)
            (13.0, 0.5928934010152285)
            (14.0, 0.49700598802395207)
            (15.0, 0.3858267716535433)
            (16.0, 0.4393939393939394)
            (17.0, 0.32608695652173914)
            (18.0, 0.3979591836734694)
            (19.0, 0.3880597014925373)
            (20.0, 0.26229508196721313)
            (21.0, 0.49056603773584906)
            (22.0, 0.36585365853658536)
            (23.0, 0.36363636363636365)
            (24.0, 0.41379310344827586)
            (25.0, 0.05102040816326531)
        };
        \end{axis}
    \end{tikzpicture}
    \hfill
    \begin{tikzpicture}
        \begin{axis}[
            width=.5\textwidth,
            height=0.3\textwidth,
            xlabel=Word length,
            ylabel=Average recall,
            mark size=1.5pt,
            legend style={font=\tiny},
            tick label style={font=\tiny},
            label style={font=\tiny},
            ytick={0.1,0.3,0.5,0.7,0.9},
            xtick={0,5,10,15,20,25},
            title={E2E TextSpotter}
        ]
        \addplot coordinates{
            (2.0, 0.13541666666666666)
            (3.0, 0.3759342301943199)
            (4.0, 0.42814877430262044)
            (5.0, 0.4585421044124778)
            (6.0, 0.46786360893180445)
            (7.0, 0.49014321819713563)
            (8.0, 0.5094901377000373)
            (9.0, 0.5093979842004903)
            (10.0, 0.4853466761972838)
            (11.0, 0.4760579064587973)
            (12.0, 0.46222887060583395)
            (13.0, 0.47381546134663344)
            (14.0, 0.41647597254004576)
            (15.0, 0.3728813559322034)
            (16.0, 0.415929203539823)
            (17.0, 0.3854748603351955)
            (18.0, 0.35772357723577236)
            (19.0, 0.3076923076923077)
            (20.0, 0.35384615384615387)
            (21.0, 0.20512820512820512)
            (22.0, 0.21428571428571427)
            (23.0, 0.2777777777777778)
            (24.0, 0.3076923076923077)
            (25.0, 0.03278688524590164)
        };
        \addplot coordinates {
            (2.0, 0.5835993663945412)
            (3.0, 0.3575187171122527)
            (4.0, 0.687397628651754)
            (5.0, 0.7154200306932675)
            (6.0, 0.7231765213638326)
            (7.0, 0.7190267905959541)
            (8.0, 0.7208708828455219)
            (9.0, 0.7049578328648096)
            (10.0, 0.704063776006222)
            (11.0, 0.6486578321440707)
            (12.0, 0.6217084136159281)
            (13.0, 0.5177664974619289)
            (14.0, 0.4411177644710579)
            (15.0, 0.30708661417322836)
            (16.0, 0.31313131313131315)
            (17.0, 0.15217391304347827)
            (18.0, 0.25510204081632654)
            (19.0, 0.2537313432835821)
            (20.0, 0.14754098360655737)
            (21.0, 0.16981132075471697)
            (22.0, 0.2682926829268293)
            (23.0, 0.3181818181818182)
            (24.0, 0.27586206896551724)
            (25.0, 0.01020408163265306)
        };
        \end{axis}
    \end{tikzpicture}
    \hfill
    \begin{tikzpicture}
        \begin{axis}[
            width=.5\textwidth,
            height=0.3\textwidth,
            xlabel=Word length,
            ylabel=Average recall,
            mark size=1.5pt,
            legend style={font=\tiny},
            tick label style={font=\tiny},
            label style={font=\tiny},
            ytick={0.1,0.3,0.5,0.7,0.9},
            xtick={0,5,10,15,20,25},
            title={oCLIP\_v2}
        ]
        \addplot coordinates{
            (2.0, 0.14583333333333334)
            (3.0, 0.3310911808669656)
            (4.0, 0.4150464919695689)
            (5.0, 0.46177468886374656)
            (6.0, 0.4772178636089318)
            (7.0, 0.5122156697556866)
            (8.0, 0.5351693338295497)
            (9.0, 0.5341868700626532)
            (10.0, 0.5378842030021443)
            (11.0, 0.5089086859688196)
            (12.0, 0.5168287210172027)
            (13.0, 0.5037406483790524)
            (14.0, 0.4759725400457666)
            (15.0, 0.43389830508474575)
            (16.0, 0.45132743362831856)
            (17.0, 0.4022346368715084)
            (18.0, 0.4146341463414634)
            (19.0, 0.375)
            (20.0, 0.4153846153846154)
            (21.0, 0.358974358974359)
            (22.0, 0.2857142857142857)
            (23.0, 0.16666666666666666)
            (24.0, 0.46153846153846156)
            (25.0, 0.09836065573770492)
        };
        \addplot coordinates {
            (2.0, 0.5281345193127818)
            (3.0, 0.3301711100278726)
            (4.0, 0.6385527441633052)
            (5.0, 0.6817908854340429)
            (6.0, 0.7038843331894692)
            (7.0, 0.6997813012575178)
            (8.0, 0.717601084615998)
            (9.0, 0.7021466905187835)
            (10.0, 0.6982306047054249)
            (11.0, 0.6591913013931363)
            (12.0, 0.636480411046885)
            (13.0, 0.5472081218274112)
            (14.0, 0.45109780439121755)
            (15.0, 0.33070866141732286)
            (16.0, 0.3333333333333333)
            (17.0, 0.2391304347826087)
            (18.0, 0.2755102040816326)
            (19.0, 0.3283582089552239)
            (20.0, 0.19672131147540983)
            (21.0, 0.3584905660377358)
            (22.0, 0.34146341463414637)
            (23.0, 0.36363636363636365)
            (24.0, 0.3793103448275862)
            (25.0, 0.030612244897959183)
        };
        \end{axis}
    \end{tikzpicture}
    \hfill
    \begin{tikzpicture}
        \begin{axis}[
            width=.5\textwidth,
            height=0.3\textwidth,
            xlabel=Word length,
            ylabel=Average recall,
            mark size=1.5pt,
            legend style={font=\tiny},
            tick label style={font=\tiny},
            label style={font=\tiny},
            ytick={0.1,0.3,0.5,0.7,0.9},
            xtick={0,5,10,15,20,25},
            title={CLOVA OCR DEER}
        ]
        \addplot coordinates{
            (2.0, 0.125)
            (3.0, 0.35575485799701045)
            (4.0, 0.4173710904480135)
            (5.0, 0.4545013738483918)
            (6.0, 0.48702474351237174)
            (7.0, 0.5016006739679865)
            (8.0, 0.5349832526981764)
            (9.0, 0.5363661127758104)
            (10.0, 0.5328806290207291)
            (11.0, 0.5244988864142539)
            (12.0, 0.5220643231114436)
            (13.0, 0.5224438902743143)
            (14.0, 0.505720823798627)
            (15.0, 0.4711864406779661)
            (16.0, 0.504424778761062)
            (17.0, 0.48044692737430167)
            (18.0, 0.4959349593495935)
            (19.0, 0.5192307692307693)
            (20.0, 0.4461538461538462)
            (21.0, 0.5128205128205128)
            (22.0, 0.2619047619047619)
            (23.0, 0.2777777777777778)
            (24.0, 0.5384615384615384)
            (25.0, 0.13114754098360656)
        };
        \addplot coordinates {
            (2.0, 0.5898135737784818)
            (3.0, 0.35356906270701316)
            (4.0, 0.680625840361814)
            (5.0, 0.7149863214786149)
            (6.0, 0.7321536469572724)
            (7.0, 0.7326954620010935)
            (8.0, 0.7407289257516548)
            (9.0, 0.7298747763864043)
            (10.0, 0.7314796811199689)
            (11.0, 0.6877336051647979)
            (12.0, 0.6814386640976237)
            (13.0, 0.5857868020304569)
            (14.0, 0.4930139720558882)
            (15.0, 0.3937007874015748)
            (16.0, 0.4393939393939394)
            (17.0, 0.32608695652173914)
            (18.0, 0.40816326530612246)
            (19.0, 0.3582089552238806)
            (20.0, 0.22950819672131148)
            (21.0, 0.49056603773584906)
            (22.0, 0.3902439024390244)
            (23.0, 0.4090909090909091)
            (24.0, 0.41379310344827586)
            (25.0, 0.04591836734693878)
        };
        \end{axis}
    \end{tikzpicture}



    \caption{Average recall for different character lengths in the End-to-End task. The red and blue lines represent the average recall for IV and OOV words, respectively.}
    \label{plot:len_e2e}
    \vspace{-0.5cm}
\end{figure}
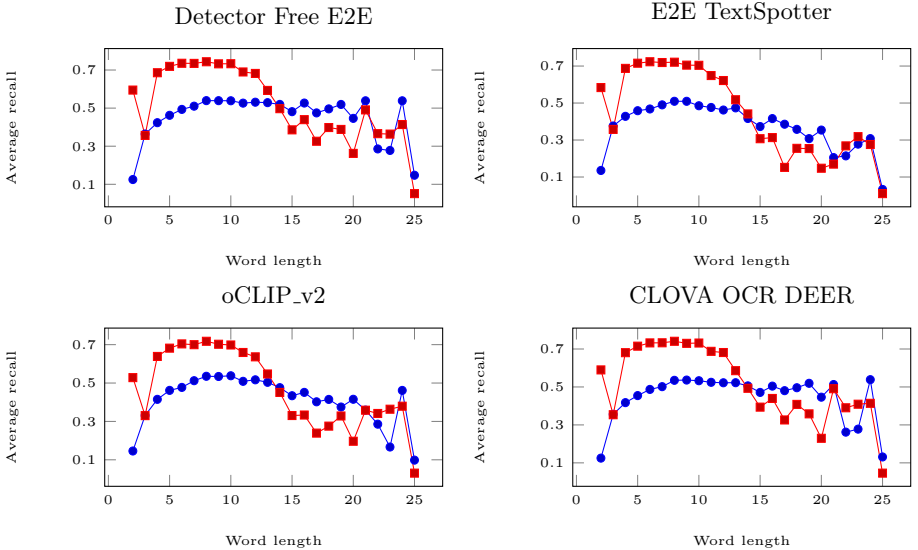

\setlength\tabcolsep{1.2pt}

\begin{figure*}[t!]
\begin{center}
\small
\begin{tabular}{p{0.5\linewidth} p{0.5\linewidth}}
    \includegraphics[width=\linewidth,height=0.75\linewidth]{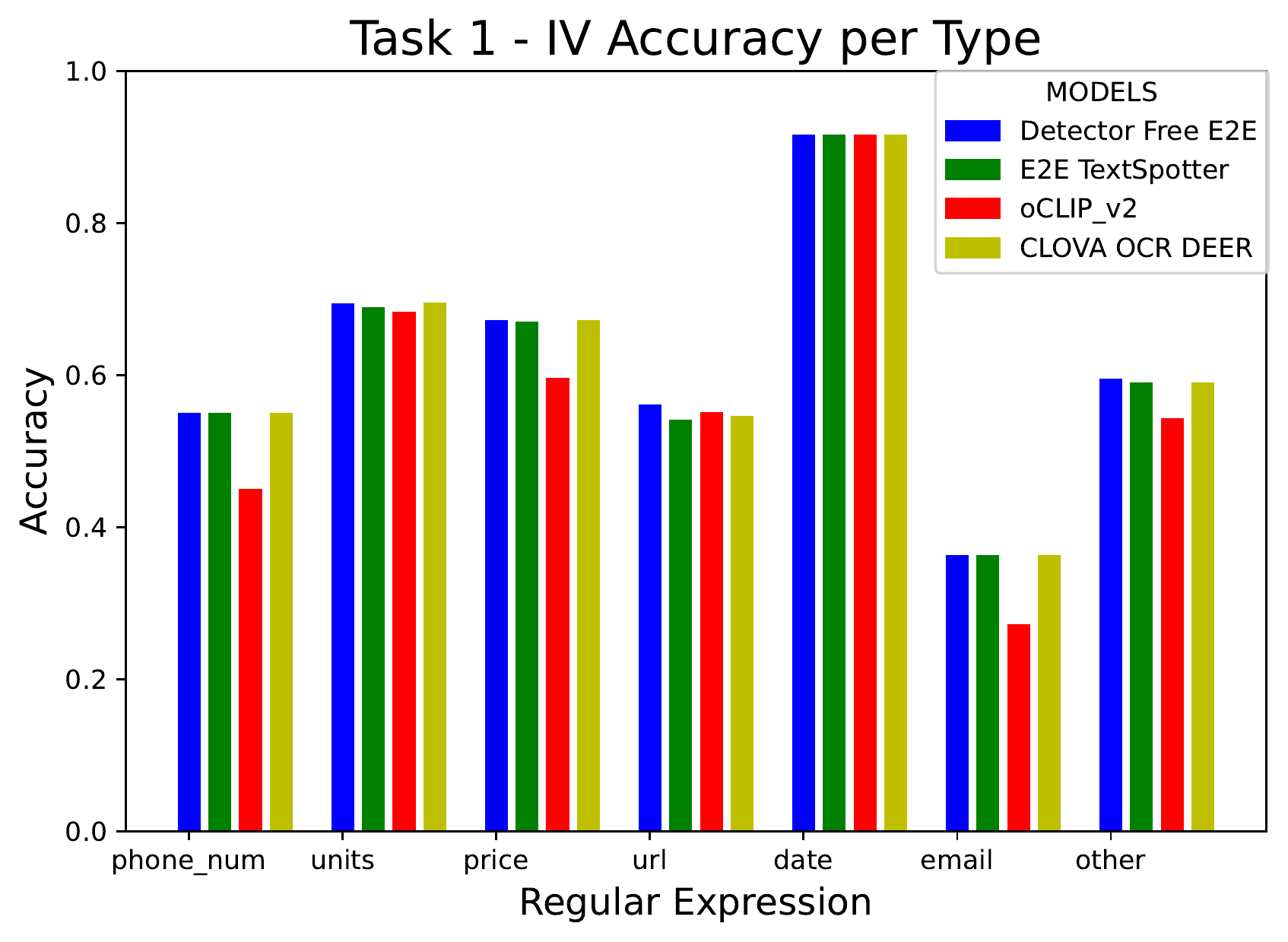}
    & 
    \includegraphics[width=\linewidth,height=0.75\linewidth]{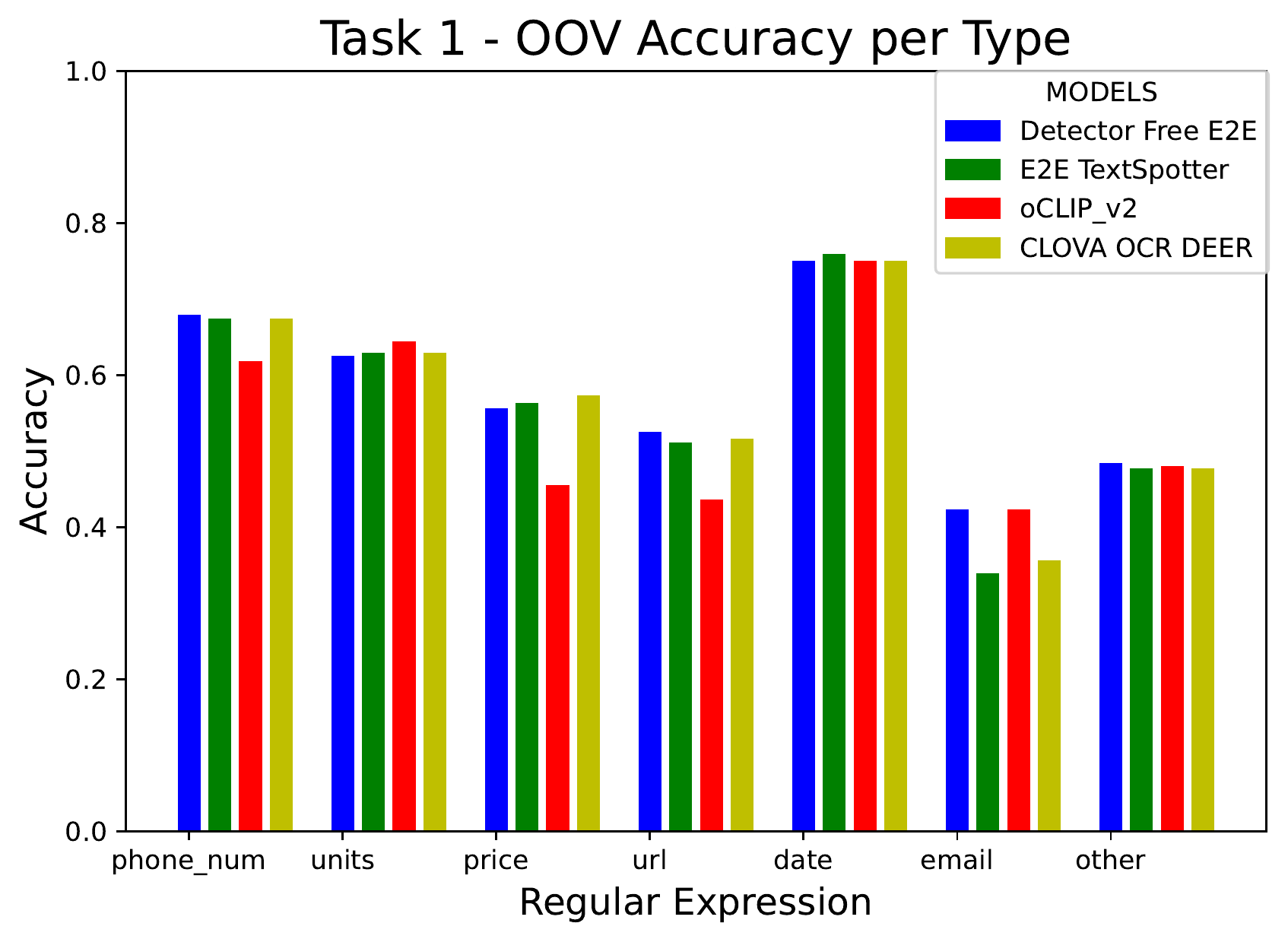}
    \\

\end{tabular}
\end{center}
\vspace{-0.8cm}
\caption{Model accuracy according to the type of scene text categorized by regular expressions in Task 1. Best viewed in color.}
\label{fig:regex_task1}
\vspace{-0.5cm}
\end{figure*}

In this section, we analyze the performance of the top submitted methods in terms of word length and different word categories. Specifically, we study whether the word length has an effect on the performance of a model and how well they perform in terms of the category of the word. 

\subsection{Task 1}

Figure \ref{plot:len_e2e} shows the performance of the top 4 methods for words of different character lengths. The metric reported is the average recall of each method for both OOV and IV words. For all methods, the results on IV words shorter than 15 characters are higher than for OOV words. The models seem to have less difficulty dealing with short in-vocabulary words, most likely as a product of vocabulary reliance. For IV words longer than 15 characters, the performance is comparable or sometimes even worse than for OOV words of the same lengths. Interestingly, results on OOV words seem to be consistent regardless of the word length, although we seem to observe fluctuations on the score for words longer than 20 characters. This could be attributed to statistical anomalies due to the low number of OOV words of this length (as seen in Figure \ref{fig:word_len_dist}, there are fewer words with of than 20 characters).

Additionally, Figure~\ref{fig:regex_task1} shows the recognition accuracy by employing an automatic categorization of words via the usage of regular expressions. In both scenarios (IV and OOV), whenever a scene text instance is mostly formed by numbers (units, prices, phone numbers) the accuracy remains uniform. We hypothesize that this outcome is a direct effect of the distribution of the training data that contains numbers. Since numbers do not follow a specific distribution as characters in a given language, scene text models are more flexible at correctly predicting numbers. Subsequently, the direct effect is found in categories where numbers are not common or absent at all, such as in emails, urls and others.


\subsection{Task 2}
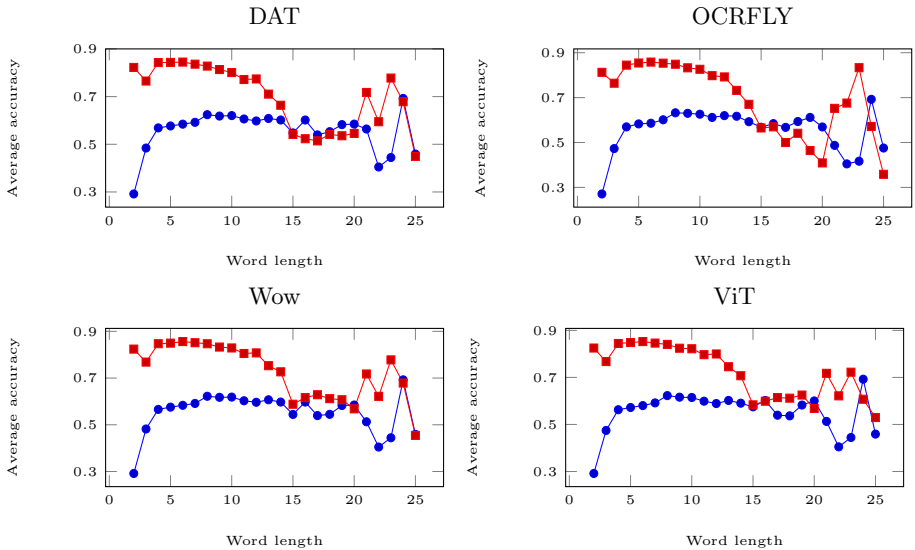
\begin{figure}[t!]
    \centering

    \begin{tikzpicture}
        \begin{axis}[
            width=.5\textwidth,
            height=0.3\textwidth,
            xlabel=Word length,
            ylabel=Average accuracy,
            mark size=1.5pt,
            legend style={font=\tiny},
            tick label style={font=\tiny},
            label style={font=\tiny},
            ytick={0.1,0.3,0.5,0.7,0.9},
            xtick={0,5,10,15,20,25},
            title={DAT}
        ]
        \addplot coordinates{
            (2.0, 0.2916666666666667)
            (3.0, 0.48423423423423423)
            (4.0, 0.568733738074588)
            (5.0, 0.576835607537362)
            (6.0, 0.5844057532172596)
            (7.0, 0.5921052631578947)
            (8.0, 0.6241372878194367)
            (9.0, 0.6183851609383524)
            (10.0, 0.6203438395415473)
            (11.0, 0.6061281337047354)
            (12.0, 0.5978915662650602)
            (13.0, 0.6082603254067585)
            (14.0, 0.6018306636155606)
            (15.0, 0.5476190476190477)
            (16.0, 0.6017699115044248)
            (17.0, 0.5393258426966292)
            (18.0, 0.5528455284552846)
            (19.0, 0.5825242718446602)
            (20.0, 0.5846153846153846)
            (21.0, 0.5641025641025641)
            (22.0, 0.40476190476190477)
            (23.0, 0.4444444444444444)
            (24.0, 0.6923076923076923)
            (25.0, 0.45901639344262296)
        };
        \addplot coordinates {
            (2.0, 0.8223024997363148)
            (3.0, 0.7653640195966731)
            (4.0, 0.8427168057462765)
            (5.0, 0.8435155412647374)
            (6.0, 0.8448157786604574)
            (7.0, 0.8360995850622407)
            (8.0, 0.8280228420094348)
            (9.0, 0.8135324435046914)
            (10.0, 0.8009630818619583)
            (11.0, 0.7716201906106601)
            (12.0, 0.7740416946872899)
            (13.0, 0.7103825136612022)
            (14.0, 0.6637744034707158)
            (15.0, 0.5416666666666666)
            (16.0, 0.5232558139534884)
            (17.0, 0.5142857142857142)
            (18.0, 0.5411764705882353)
            (19.0, 0.5357142857142857)
            (20.0, 0.5454545454545454)
            (21.0, 0.717391304347826)
            (22.0, 0.5945945945945946)
            (23.0, 0.7777777777777778)
            (24.0, 0.6785714285714286)
            (25.0, 0.44919786096256686)
        };
        \end{axis}
    \end{tikzpicture}
    \hfill
    \begin{tikzpicture}
        \begin{axis}[
            width=.5\textwidth,
            height=0.3\textwidth,
            xlabel=Word length,
            ylabel=Average accuracy,
            mark size=1.5pt,
            legend style={font=\tiny},
            tick label style={font=\tiny},
            label style={font=\tiny},
            ytick={0.1,0.3,0.5,0.7,0.9},
            xtick={0,5,10,15,20,25},
            title={OCRFLY}
        ]
        \addplot coordinates{
            (2.0, 0.2708333333333333)
            (3.0, 0.47297297297297297)
            (4.0, 0.5696010407632264)
            (5.0, 0.5828460038986355)
            (6.0, 0.5857683573050719)
            (7.0, 0.6012145748987854)
            (8.0, 0.6323447118074986)
            (9.0, 0.630114566284779)
            (10.0, 0.6260744985673352)
            (11.0, 0.611699164345404)
            (12.0, 0.6197289156626506)
            (13.0, 0.6170212765957447)
            (14.0, 0.5926773455377574)
            (15.0, 0.5680272108843537)
            (16.0, 0.584070796460177)
            (17.0, 0.5674157303370787)
            (18.0, 0.5934959349593496)
            (19.0, 0.6116504854368932)
            (20.0, 0.5692307692307692)
            (21.0, 0.48717948717948717)
            (22.0, 0.40476190476190477)
            (23.0, 0.4166666666666667)
            (24.0, 0.6923076923076923)
            (25.0, 0.47540983606557374)
        };
        \addplot coordinates {
            (2.0, 0.8127043560805822)
            (3.0, 0.7642018913068247)
            (4.0, 0.8446973698109221)
            (5.0, 0.8541264737406217)
            (6.0, 0.8582842532986349)
            (7.0, 0.8535039188566159)
            (8.0, 0.848713067946702)
            (9.0, 0.8326945949517642)
            (10.0, 0.8262439807383628)
            (11.0, 0.7977409106953759)
            (12.0, 0.7921990585070612)
            (13.0, 0.73224043715847)
            (14.0, 0.6702819956616052)
            (15.0, 0.5648148148148148)
            (16.0, 0.5697674418604651)
            (17.0, 0.5)
            (18.0, 0.5411764705882353)
            (19.0, 0.4642857142857143)
            (20.0, 0.4090909090909091)
            (21.0, 0.6521739130434783)
            (22.0, 0.6756756756756757)
            (23.0, 0.8333333333333334)
            (24.0, 0.5714285714285714)
            (25.0, 0.3582887700534759)
        };
        \end{axis}
    \end{tikzpicture}
    \hfill
    \begin{tikzpicture}
        \begin{axis}[
            width=.5\textwidth,
            height=0.3\textwidth,
            xlabel=Word length,
            ylabel=Average accuracy,
            mark size=1.5pt,
            legend style={font=\tiny},
            tick label style={font=\tiny},
            label style={font=\tiny},
            ytick={0.1,0.3,0.5,0.7,0.9},
            xtick={0,5,10,15,20,25},
            title={Wow}
        ]
        \addplot coordinates{
            (2.0, 0.2916666666666667)
            (3.0, 0.481981981981982)
            (4.0, 0.5659150043365134)
            (5.0, 0.5755360623781677)
            (6.0, 0.5833459500378501)
            (7.0, 0.5912618083670715)
            (8.0, 0.6220854318224212)
            (9.0, 0.6175668303327878)
            (10.0, 0.6185530085959885)
            (11.0, 0.6027855153203343)
            (12.0, 0.5963855421686747)
            (13.0, 0.6070087609511889)
            (14.0, 0.597254004576659)
            (15.0, 0.54421768707483)
            (16.0, 0.5973451327433629)
            (17.0, 0.5393258426966292)
            (18.0, 0.5447154471544715)
            (19.0, 0.5825242718446602)
            (20.0, 0.5846153846153846)
            (21.0, 0.5128205128205128)
            (22.0, 0.40476190476190477)
            (23.0, 0.4444444444444444)
            (24.0, 0.6923076923076923)
            (25.0, 0.45901639344262296)
        };
        \addplot coordinates {
            (2.0, 0.823990085434026)
            (3.0, 0.7683490942235388)
            (4.0, 0.8469684166050491)
            (5.0, 0.8494819578420865)
            (6.0, 0.8560927726795416)
            (7.0, 0.851313969571231)
            (8.0, 0.8468923280642224)
            (9.0, 0.8326945949517642)
            (10.0, 0.8288523274478331)
            (11.0, 0.8051535474761736)
            (12.0, 0.8076664425016813)
            (13.0, 0.7530054644808744)
            (14.0, 0.7266811279826464)
            (15.0, 0.5879629629629629)
            (16.0, 0.6162790697674418)
            (17.0, 0.6285714285714286)
            (18.0, 0.611764705882353)
            (19.0, 0.6071428571428571)
            (20.0, 0.5681818181818182)
            (21.0, 0.717391304347826)
            (22.0, 0.6216216216216216)
            (23.0, 0.7777777777777778)
            (24.0, 0.6785714285714286)
            (25.0, 0.45454545454545453)
        };
        \end{axis}
    \end{tikzpicture}
    \hfill
    \begin{tikzpicture}
        \begin{axis}[
            width=.5\textwidth,
            height=0.3\textwidth,
            xlabel=Word length,
            ylabel=Average accuracy,
            mark size=1.5pt,
            legend style={font=\tiny},
            tick label style={font=\tiny},
            label style={font=\tiny},
            ytick={0.1,0.3,0.5,0.7,0.9},
            xtick={0,5,10,15,20,25},
            title={ViT}
        ]
        \addplot coordinates{
            (2.0, 0.2916666666666667)
            (3.0, 0.4744744744744745)
            (4.0, 0.5626626192541196)
            (5.0, 0.5721247563352827)
            (6.0, 0.5801665404996215)
            (7.0, 0.5915991902834008)
            (8.0, 0.6226450289125163)
            (9.0, 0.6159301691216584)
            (10.0, 0.6146131805157593)
            (11.0, 0.5988857938718662)
            (12.0, 0.588855421686747)
            (13.0, 0.6020025031289111)
            (14.0, 0.5903890160183066)
            (15.0, 0.5748299319727891)
            (16.0, 0.6017699115044248)
            (17.0, 0.5393258426966292)
            (18.0, 0.5365853658536586)
            (19.0, 0.5825242718446602)
            (20.0, 0.6)
            (21.0, 0.5128205128205128)
            (22.0, 0.40476190476190477)
            (23.0, 0.4444444444444444)
            (24.0, 0.6923076923076923)
            (25.0, 0.45901639344262296)
        };
        \addplot coordinates {
            (2.0, 0.8249129838624618)
            (3.0, 0.7674148342258175)
            (4.0, 0.8441956269145453)
            (5.0, 0.8481957842086459)
            (6.0, 0.8522576815961284)
            (7.0, 0.845954356846473)
            (8.0, 0.840271455764297)
            (9.0, 0.8241046649927316)
            (10.0, 0.8218298555377207)
            (11.0, 0.7966819625838334)
            (12.0, 0.7995965030262273)
            (13.0, 0.7453551912568306)
            (14.0, 0.7071583514099783)
            (15.0, 0.5833333333333334)
            (16.0, 0.5988372093023255)
            (17.0, 0.6142857142857143)
            (18.0, 0.611764705882353)
            (19.0, 0.625)
            (20.0, 0.5681818181818182)
            (21.0, 0.717391304347826)
            (22.0, 0.6216216216216216)
            (23.0, 0.7222222222222222)
            (24.0, 0.6071428571428571)
            (25.0, 0.5294117647058824)
        };
        \end{axis}
    \end{tikzpicture}
    \hfill

    \caption{Average precision for different character lengths in the Cropped Word Recognition task. The red and blue lines represent the average recall for IV and OOV words, respectively.}
    \label{plot:len_crops}
    \vspace{-.7cm}
\end{figure}

\setlength\tabcolsep{1.2pt}

\begin{figure*}[t!]
\begin{center}
\small
\begin{tabular}{p{0.5\linewidth} p{0.5\linewidth}}
    \includegraphics[width=\linewidth,height=0.75\linewidth]{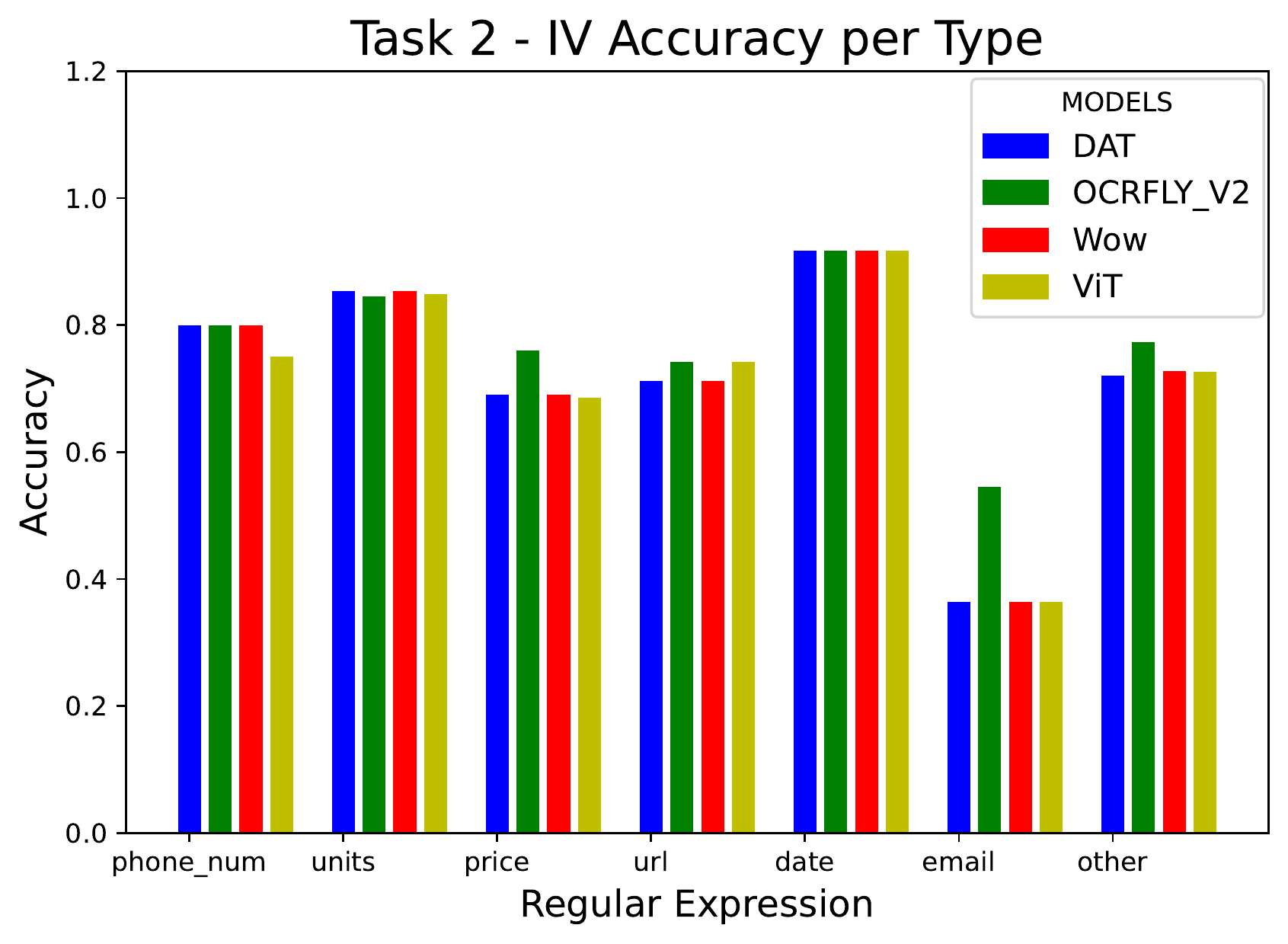} &
    \includegraphics[width=\linewidth,height=0.75\linewidth]{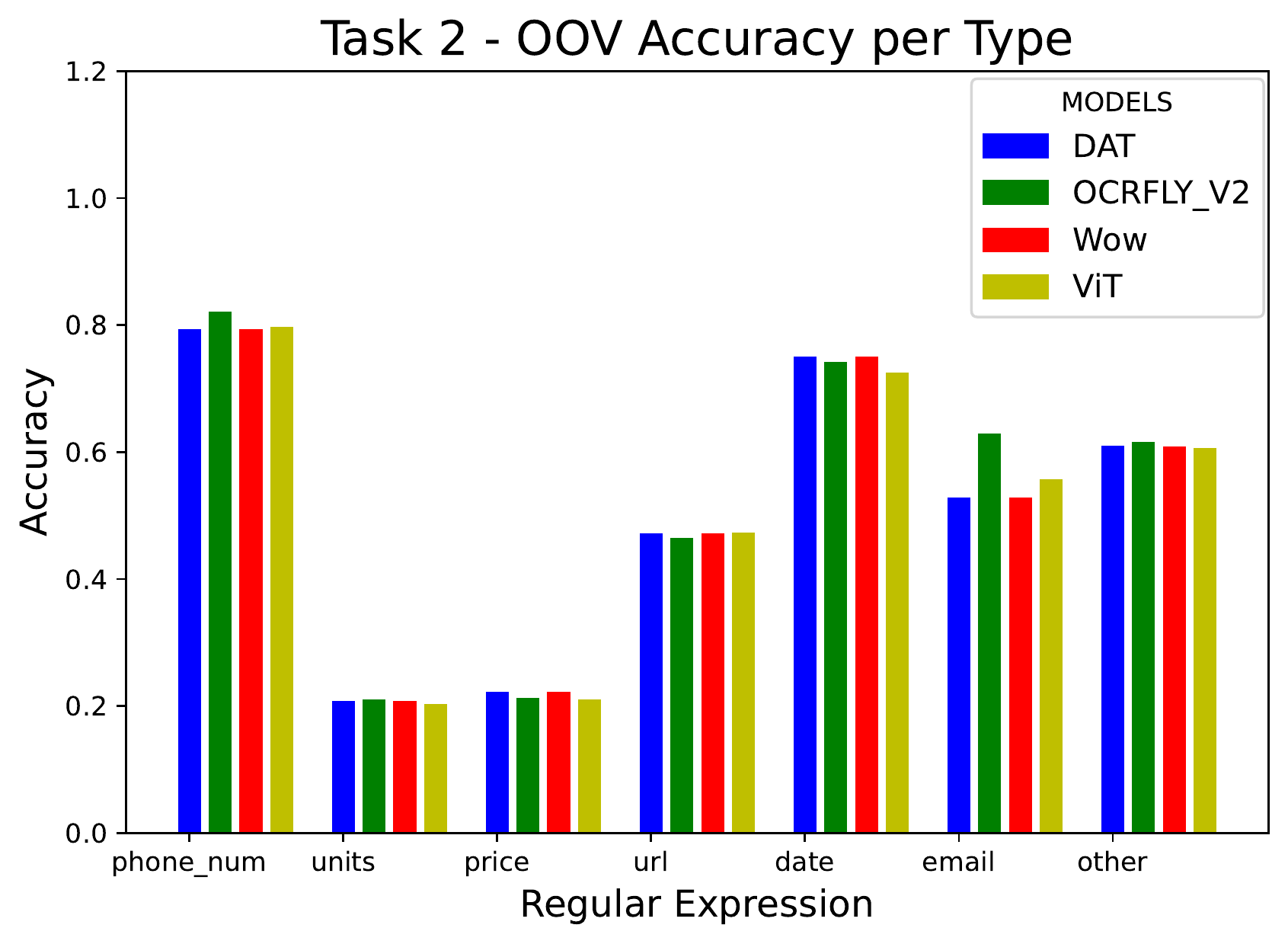} 
    \\

\end{tabular}
\end{center}
\vspace{-0.8cm}
\caption{Model accuracy according to the type of scene text categorized by regular expressions in Task 2. Best viewed in color.}
\label{fig:regex_task2}
\vspace{-0.4cm}
\end{figure*}
Figure \ref{plot:len_e2e} shows the performance of the top 4 methods for words of different lengths of characters. In this case we feature the average precision (correctly recognized words) of the top 4 submitted methods, for both OOV and IV. We observe a similar pattern as in the End-to-End task, the models seem to perform better on IV words of character length of 15 or less. The results on OOV words also seem to remain consistent for different character lengths. Like we have observed in Task 1, performance on OOV words of more than 15 characters is similar or superior to the IV of the same length, which suggest that OCR systems have trouble with longer sequences, regardless of whether they are in vocabulary or not.
Similarly to the previous subsection, we show in Figure~\ref{fig:regex_task2} the performance of the top 4 models on different word categories. Since in Task 2, no detection is involved, we observe a slightly different behaviour compared to the previous task. Whenever solely numbers are contained in a cropped image, the accuracy in IV and OOV remains similar, as in the case of phone numbers. However, if numbers and characters are expected to be found in a cropped word, the gap in performance is very large in the rest of the categories, except for emails. Even though the performance of all models is very close in task 2 ( Table~\ref{tab:recog_performance}), we observe that the winning entry, OCRFLY, gets an edge on prices, emails and other categories in IV and on phone numbers and emails in OOV words.

\section{Conclusion and Future Work}
In this work, we introduce a new task called Out-Of-Vocabulary Challenge, in which end-to-end text recognition and cropped scene text recognition were the two challenges that made up the competition.
In order to cover a wide range of data distributions, the competition creates a collection of open scene text datasets that include 326K images and 4.8M scene text instances.
Surprisingly, state-of-the-art models exhibit a considerable performance discrepancy on the OOV task. This is especially apparent in the performance gap between in-vocabulary and out-of-vocabulary words.
We come to the conclusion that in order to create scene text models that produce more reliable and generalized predictions, the OOV dataset suggested in this challenge would be a crucial area to investigate forward. 


\clearpage
%
%
\bibliographystyle{splncs04}
\bibliography{egbib}
\end{document}